%% file: main.tex
\documentclass[conference]{IEEEtran}
\usepackage{times}

\usepackage{multicol}
\usepackage{multirow}
\usepackage{graphicx}
\usepackage{algorithmic}
\usepackage{hhline}
\usepackage{algorithm}
\usepackage{amsthm}
\usepackage{amssymb}
\usepackage{amsmath}
\usepackage{dsfont}
\usepackage{booktabs}
\usepackage{wrapfig}
\usepackage[font=small, labelfont=bf]{caption}
\usepackage{tabularx}
\usepackage{soul}
\usepackage{stfloats}
\usepackage{float}

\usepackage{graphicx}
\usepackage{subcaption}
\usepackage{booktabs}
\usepackage{amsfonts}
\usepackage{xspace}
\usepackage[T1]{fontenc}
\usepackage{comment}
\usepackage{enumitem}
\usepackage{wrapfig}
\usepackage{listings}
\usepackage{float}
\usepackage{footnote}

\newcommand{\BAIR}{Office~}
\usepackage[square, numbers, sort&compress]{natbib}
\usepackage{natbib} 
\usepackage{bbm}
\usepackage[dvipsnames]{xcolor}

\usepackage[bookmarks=true]{hyperref}
\hypersetup{
    colorlinks=true,
    linkcolor=orange,
    filecolor=orange,      
    urlcolor=orange,
    citecolor=orange,
}

\usepackage[capitalise, nameinlink]{cleveref}

\begin{document}

\title{Pushing the Limits of Cross-Embodiment Learning for Manipulation and Navigation}


\author{Jonathan Yang\IEEEauthorrefmark{1}, Catherine Glossop\IEEEauthorrefmark{2}, Arjun Bhorkar\IEEEauthorrefmark{2}, Dhruv Shah\IEEEauthorrefmark{2}, Quan Vuong\IEEEauthorrefmark{3}, \\ Chelsea Finn\IEEEauthorrefmark{1}, Dorsa Sadigh\IEEEauthorrefmark{1}, Sergey Levine\IEEEauthorrefmark{2} \\
\IEEEauthorblockA{\IEEEauthorrefmark{1} Stanford University}%
\IEEEauthorblockA{\IEEEauthorrefmark{2} University of California, Berkeley}%
\IEEEauthorblockA{\IEEEauthorrefmark{3} Google Deepmind}}

\maketitle
\mbox{}
\vspace{-1.1cm}
\begin{figure}[H]
\begin{minipage}{\textwidth}\includegraphics[width=\textwidth]{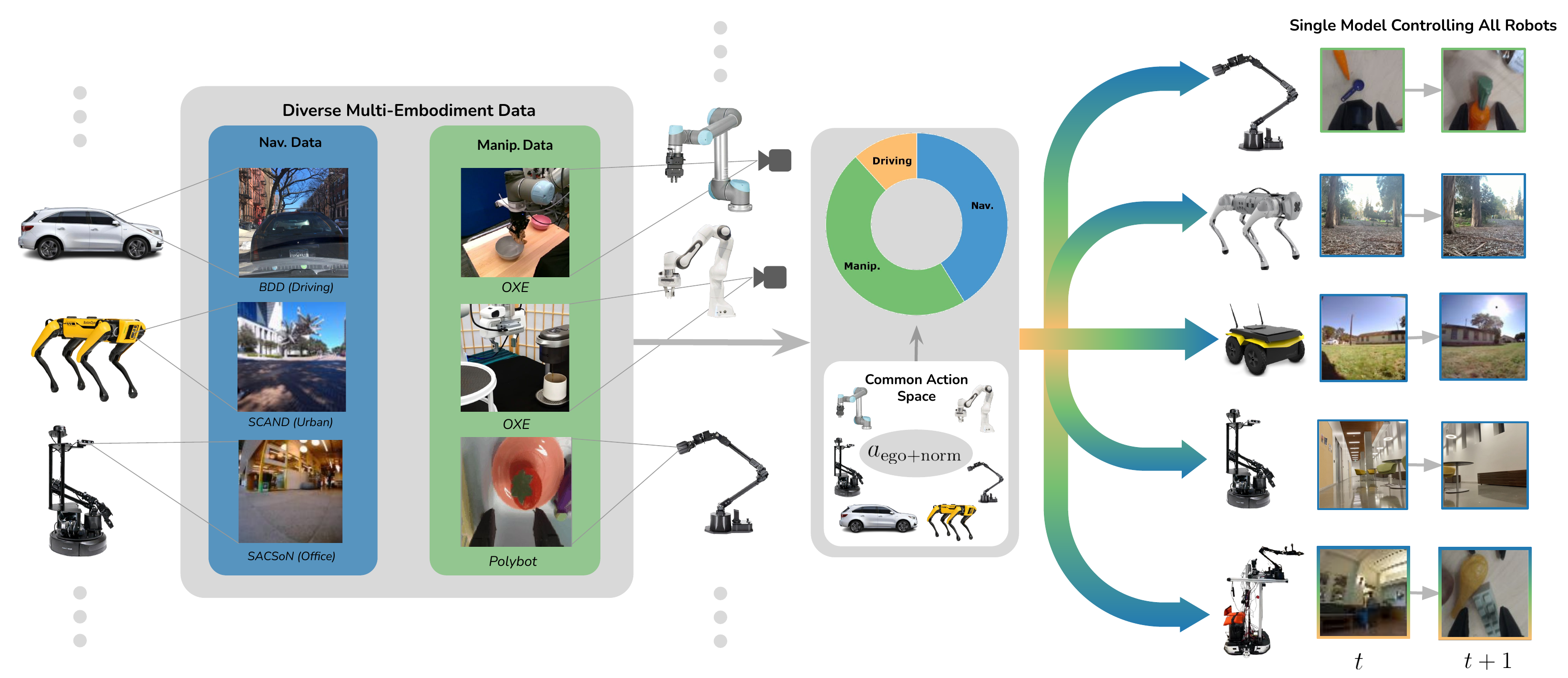}
\caption{\textbf{Heterogeneous cross-embodiment learning.} We test the limits of cross-embodiment learning by training a \emph{single goal-conditioned policy} across 18 manipulation, navigation, and driving datasets. Our policy can control a variety of manipulators, wheeled, and legged robots, as well as novel embodiments such as drones and mobile manipulators, in challenging real-world environments.}
\end{minipage}
    \label{fig:teaser}
\end{figure}

\begin{abstract}
Recent years in robotics and imitation learning have shown remarkable progress in training large-scale foundation models by leveraging data across a multitude of embodiments. The success of such policies might lead us to wonder: just how diverse can the robots in the training set be while still facilitating positive transfer? In this work, we study this question in the context of heterogeneous embodiments, examining how even seemingly very different domains, such as robotic navigation and manipulation, can provide benefits when included in the training data for the same model. We train a single goal-conditioned policy that is capable of controlling robotic arms, quadcopters, quadrupeds, and mobile bases.
We then investigate the extent to which transfer can occur across navigation and manipulation on these embodiments by framing them as a single goal-reaching task. We find that co-training with navigation data can enhance robustness and performance in goal-conditioned manipulation with a wrist-mounted camera. We then deploy our policy trained only from navigation-only and static manipulation-only data on a mobile manipulator, showing that it can control a novel embodiment in a zero-shot manner. These results provide evidence that large-scale robotic policies can benefit from data collected across various embodiments. Further information and robot videos can be found on our project \href{http://extreme-cross-embodiment.github.io}{website.\footnote{https://extreme-cross-embodiment.github.io}}
\end{abstract}



\mbox{}
\vspace{8.2cm}
\input{text/1-introduction}
\input{text/2-relatedwork}

\input{text/3-preliminaries}
\input{text/4-method}
\input{text/5-evaluation}

\section{Conclusion} 
\label{sec:conclusion}

In this paper, we analyze the effect of dataset diversity in cross-embodiment learning by blurring the boundary between navigation and manipulation. We hypothesize that projecting the different robotic tasks into a unified goal-reaching framework can lead to improved transfer of learned behaviors across embodiments, as well as to novel embodiments. We train the first \emph{heterogeneous} cross-embodiment policy capable of controlling a variety of diverse robots --- robotic arms, wheeled and legged mobile platforms, drones, and mobile manipulators --- in diverse real-world environments. We conduct over 1000 experiments to empirically characterize the effects of dataset size and variability, model size, and architecture choices. Our experiments reveal that policies co-trained with all manipulation and mobile data demonstrate an average of $20\%$ improvement over 5 different manipulation tasks than training with manipulation data alone, and a $5-7\%$ improvement over 4 different navigation platforms.
We believe such results are not merely quirks of the chosen datasets, but signs that there exists valuable information to transfer across seemingly different robot embodiments. Generalist manipulation agents would benefit from the large perceptual diversity and rich spatial relationships captured by navigation datasets, and generalist navigation agents would benefit from more the rich object-centric interactions present in manipulation datasets.

Our methodology does have a number of limitations. First, although we show that our policies can control both manipulators and mobile robots, our framework does not support systems that require controlling varying degrees of freedom. For example, although we show control of a quadrupedal robot, we are still controlling it at the level of overall heading rather than at the individual joints. Some robots, such as multi-fingered hands, do not readily support such abstraction, and extending our framework to handle varying numbers of degrees of freedom is an exciting direction for future. Second, all of our results focus on goal-conditioned policies that are tasked with a goal image. This modality is pragmatic and easy to evaluate, but not necessarily the most convenient for human users. Other task modalities, such as language, might be more useful in practice, and extending our framework to support this would also be valuable. These future improvements would make cross-embodiment training even more useful, and we hope that our work represents a step toward achieving greater synergy between robotic embodiments in the future, towards the goal of a true ``robot foundation model'' that can leverage data from all robots and control any robot out of the box.

\section*{Acknowledgments}
This research was supported by ONR grants N00014-22-1-2621 and N00014-22-1-2293, ARL DCIST CRA W911NF-17-2-0181, and NSF IIS-2150826, Ford, and Volkswagen. The authors would like to thank Pete Florence, Laura Smith, and Colin Li, for their helpful feedback on the paper. In addition, the authors would like to thank the IRIS, ILIAD, and RAIL labs for the numerous discussions about training generalist robotic policies and cross-embodiment learning.

\section{Contributions}
\noindent \textbf{Jonathan Yang:} Led model development and tuning, wrote dataloader/data processing, wrote manipulation controllers, ran manipulation evaluations, ran experiments on the mobile manipulator.\\
\textbf{Catherine Glossop:} Tuned cross-embodiment models, implemented and trained ablations models, evaluated runs for the locobot, created plots/videos for the paper.  \\
\textbf{Arjun Bhorkar:} Worked on data processing and model training for navigation, created navigation pipelines and ran evaluations for the DJI Tello, Clearpath Jackal and Unitree Go1. \\
\textbf{Dhruv Shah:} Provided guidance for the project and technical report, helped resolve issues with navigation. \\
\textbf{Quan Vuong:} Provided guidance for the project, helped with utilizing and dataloading from RT-X. \\
\textbf{Chelsea Finn, Dorsa Sadigh, Sergey Levine:} Provided guidance for the project and the technical report.


\bibliographystyle{IEEEtran}
\bibliography{references.bib}

\newpage
\input{text/7-appendix}

\end{document}

%% file: text/1-introduction.tex
\section{Introduction}

The advent of large-scale foundation models in machine learning has enabled harnessing diverse datasets to enhance sample efficiency, improve generalization, and facilitate transfer to novel domains~\cite{bommasani2022opportunities}. Recent years in robotics have seen an acceleration in the collection and consolidation of large-scale datasets in the hopes of obtaining similar benefits. These datasets have contained demonstrations spanning many scenes, observations, viewpoints, tasks, and embodiments in a wide range of robotics domains such as manipulation \cite{dasari2020robonet, fang2023rh20t, embodimentcollaboration2023open}, navigation \cite{shah2023gnm, tsai2023multimodal}, autonomous driving \cite{wang2023drive}, and others \cite{lin2023motionx}. However, these prior works typically restrict their investigations to sets of similar embodiments -- e.g., arms with parallel jaw grippers. In contrast, the most successful large-scale foundation models are typically trained on highly heterogeneous data, such as large text corpora mined from the web.
This raises the question: what degree of embodiment diversity can we include when training broadly capable ``generalist'' robot policies? 

We study this problem in the context of heterogeneous embodiments, aiming to understand whether large-scale policies can benefit from data across navigation and manipulation. Enabling this transfer of knowledge can eliminate the need to recollect datasets containing information present in one domain but not the other. For example, navigation data can help manipulators understand spatial relationships between different poses. Similarly, manipulation data can help navigators with object-centric reasoning. This is particularly crucial for mobile manipulators, which need both the mobile base and the robotic arm to approach certain objects.

Why might we hope for positive transfer across navigation and manipulation? While these domains seemingly differ significantly in terms of hardware, observations, and action representations, they contain many similar sensorimotor principles. For example, both domains require the learned robot policy to have an understanding of collisions and geometry. Both domains also require the agent to perform some form of visual servoing. In manipulation, the robot analyzes its observation to determine the position of its end-effector with respect to a target object, and then moves towards it. Similarly, in visual navigation, the robot examines the spatial relationship between its current location and goal, as inferred from image observations, and determines how to move toward the goal. If the manipulation task uses a wrist-mounted camera and the navigation task uses a forward-facing camera, both embodiments have the same equivariance between pose changes and camera observations -- i.e., moving ``left'' with respect to the image will transform the observations in egocentric manipulation and navigation in a similar manner.



In this paper, we empirically investigate the benefits of including navigation data for robotic manipulation, and vice versa. We present, to our knowledge, the first results demonstrating a large-scale policy trained jointly on navigation and manipulation data from many different robots, showing that such a policy can control robotic arms, drones, quadrupeds, mobile bases, and mobile manipulators.
We then demonstrate that a co-trained policy can achieve a $20\%$
higher success rate over a manipulation-only policy. Interestingly, the same co-trained policy achieves a $5-7\%$ improvement over a navigation-only policy on 4 different robots. This suggests that robotic agents can benefit from data collected across significantly different embodiments. We then characterize which datasets are most useful for manipulation and demonstrate that navigation data helps the policy learn embeddings that are more informative of distance to the goal in novel manipulation environments. We finally show that our policy can generalize to two new robots: a mobile manipulator and a quadrotor, without any data specific to these embodiments. 
While the particular training methodology and model architecture are based on prior techniques, the empirical findings are a novel contribution of our work, demonstrating for the first time that navigation data can provide quantifiable benefits for robotic manipulation in the cross-embodied policy learning setting.

%% file: text/2-relatedwork.tex
\section{Related Work}

Traditionally, robotic learning has involved training policies using datasets specifically gathered for each robot and its designated task. However, the substantial cost of data collection and the ensuing lack of diversity in these datasets have resulted in policies with notably limited generalization ability. To alleviate this issue, several prior works have investigated cross-embodiment transfer at a small scale to enable the reuse of robotic datasets. In this section, we will first describe the body of work focused on cross-embodiment transfer at a small scale. We will then discuss several efforts in collecting large robotics datasets and training policies in these settings. 

\smallskip \noindent \textbf{Cross-embodiment transfer.} Prior works on cross-embodiment transfer have typically focused on transferring to novel robot parametrizations in simulation~\cite{yu2017preparing, chen2019hardware, pmlr-v180-you22a},
novel morphologies in the real world \cite{hu2022know, salhotra2023bridging, yang2023polybot}, and via sim-to-real transfer \cite{christiano2016transfer, sadeghi2017sim2real, Peng_2018, zhang2021cycleconsistency}. By conditioning policies on embodiment \cite{devin2016learning, wang2016nervenet, yu2017preparing, chen2019hardware, huang2020policy, ghadirzadeh2021bayesian, hirose2022exaug, salhotra2023bridging, attarian2023geometry}, unifying action abstractions \cite{Loquercio_2018, martin2019variable, chang2020semantic, Shao_2020, kang2021hierarchically, baul2023affordances, shah2023gnm, shah2023vint}, and using domain adaptation \cite{ganin2016domainadversarial, bousmalis2017using, gupta2017learning, fang2018multitask, kim2020learning, zhu2020unpaired, zhang2021cycleconsistency, you2022crossdomain, yang2023polybot}, these works showed that robotic policies can generalize to new tasks and environments on different embodiments. Due to the large diversity of robot hardware, enabling the capability to learn from one robot hardware to control another is crucial to leveraging the robot data that is currently available. Our work specifically focuses on the problem of transferring knowledge between real-world robotic policies across a heterogeneous set of embodiments, or embodiments spanning navigation and manipulation with large amounts of variety in hardware.

\smallskip \noindent \textbf{Large-scale robotic datasets and policies.} While these smaller-scale projects have demonstrated great success in facilitating multi-robot transfer, it has become clear that training policies that apply to a large variety of embodiments and domains would require learning from large and diverse datasets. To address this issue, researchers have created real-world robotic datasets for manipulation \cite{sharma2018multiple, mandlekar2018roboturk, dasari2020robonet, young2020visual, jang2021bcz, ebert2021bridge, fang2023rh20t, zhao2023learning}, navigation \cite{hirose2019deep, shah2023gnm, karnan2022socially}, and autonomous driving \cite{cordts2016cityscapes, 6248074, geyer2020a2d2, Huang_2020, yu2020bdd100k, Ettinger_2021_ICCV}. The increased availability of open-source robotic datasets has enabled training large-scale robotic foundation models that leverage data across many embodiments \cite{reed2022generalist, shah2023gnm, bousmalis2023robocat, embodimentcollaboration2023open, octo_2023, hu2023gaia1}. These foundation models include object tracking models \cite{goodwin2022zeroshot, zhu2023viola, zhu2023learning}, representation learning models \cite{bonatti2022pact, karamcheti2023languagedriven}, and predictive world models \cite{dasari2020robonet, du2023learning, xian2023generalist, yang2024learning}.

A number of robotic foundation models referred to in recent work as Generalist Robot Policies (GRPs)
have been trained on large, diverse datasets to directly predict low-level robot actions from image observations \cite{brohan2023rt1, brohan2023rt2, shah2023gnm,  embodimentcollaboration2023open, octo_2023, shah2023vint}. These foundation models have typically been co-trained \cite{embodimentcollaboration2023open} or pre-trained \cite{reed2022generalist, octo_2023} with data across multiple embodiments. GRPs have demonstrated the ability to fit a broad range of embodiments and significantly enhance their performance in new tasks and environments \cite{embodimentcollaboration2023open, octo_2023}. While these models have previously been trained on datasets exclusively consisting of manipulation, navigation, or driving data, we propose to train a single model that can benefit from robotic data across these domains. We then investigate the extent to which this generalization can occur across these significantly different embodiments.

%% file: text/3-preliminaries.tex
\section{Preliminaries}
We will study heterogeneous cross-embodiment robotic learning in the context of goal-conditioned imitation learning and training policies to reach visually indicated goals from data. Let us define datasets as $D_{e} := \{\tau_{1}, \tau_{2}, \ldots, \tau_{k}\}$ consisting of $k$ demonstrations for embodiment $e$. Each trajectory $\tau \in D_{e_{m}}$ consists of a sequence of observations (images) and actions. That is, $\tau := \{o_{0}, a_{0}, o_{1}, a_{1}, \ldots \}$. The objective of goal-conditioned imitation learning is to train a policy $\pi(a|o, o_{g})$ to output actions that control a particular embodiment given the current and goal observations.

\smallskip \noindent \textbf{Goal-conditioned manipulation.}
In goal-conditioned manipulation, the policy must learn to output a sequence of actions that are converted to joint velocities and given to a lower-level controller. Manipulation datasets $D_{e_m}$ ($e_m$ referring to manipulation for the embodiment) typically consist of teleoperated demonstrations from a remote controller, VR headset, or a haptic device. These different modalities can lead to demonstrations that contain many different choices for actions such as absolute and delta Cartesian control, absolute and delta joint control, or operational space control. Even with a similar controller, differences in coordinate frame and gains can cause discrepancies in action interpretation between robots.

\smallskip \noindent \textbf{Visual navigation.}
The objective of visual robotic navigation is to direct a robotic agent to move to a goal $g \in G$ while avoiding obstacles. The robot is not given any ground-truth localization information, GPS readings, or semantic maps requiring it to output a series of waypoints or velocities given only its observation history and goal image. In addition, the agent predicts a distance function $d(\cdot|o_{t-k: t}, o_{g})$ to determine the distance between its current observation and its goal. During evaluation time, the robot is given a topological map $\mathcal{M}$, which is a sequence of image subgoals. The agent must first determine a feasible subgoal $o_{g} \in \mathcal{M}$ and then determine how to move to this subgoal using a local policy. The subgoal is determined by querying the distance function on all of the goal images in the topological map, then determining the closest image to the robot.


%% file: text/4-method.tex
\section{Heterogeneous Cross-Embodiment Learning}
In this work, we study cross-embodiment robotic learning with embodiments that include navigation platforms and robotic arms. We refer to this as heterogeneous cross-embodiment, to distinguish it from earlier works that studied cross-embodiment with data from similar robots and near-identical action spaces~\citep{dasari2020robonet, hu2022know}. Given manipulation datasets $D_{e_{m, \cdot}}$ and navigation datasets $D_{e_{n, \cdot}}$, we would like to learn a \textit{single} policy $D_{e_{m, 1}} \cup D_{e_{m, 2}} \cup \ldots \cup D_{e_{n, 1}} \cup D_{e_{n, 2}}$ that can control robots in both domains. To solve this problem, we train a goal-conditioned policy $\pi(a | o, o_{g})$ that outputs $k$ actions into the future given a context of $c$ observations. While we could simply train a single policy across all of the navigation and manipulation datasets to output action labels that match each specific dataset (using padding or sequence models in case dimensionalities do not match), we propose a unified action and observation representation that is specifically designed for our cross-embodied training setting in the following sections.


\subsection{Manipulation and Navigation as Unified Goal-Reaching}

\begin{figure}
\vspace*{-1em}
\includegraphics[width=0.5\textwidth]{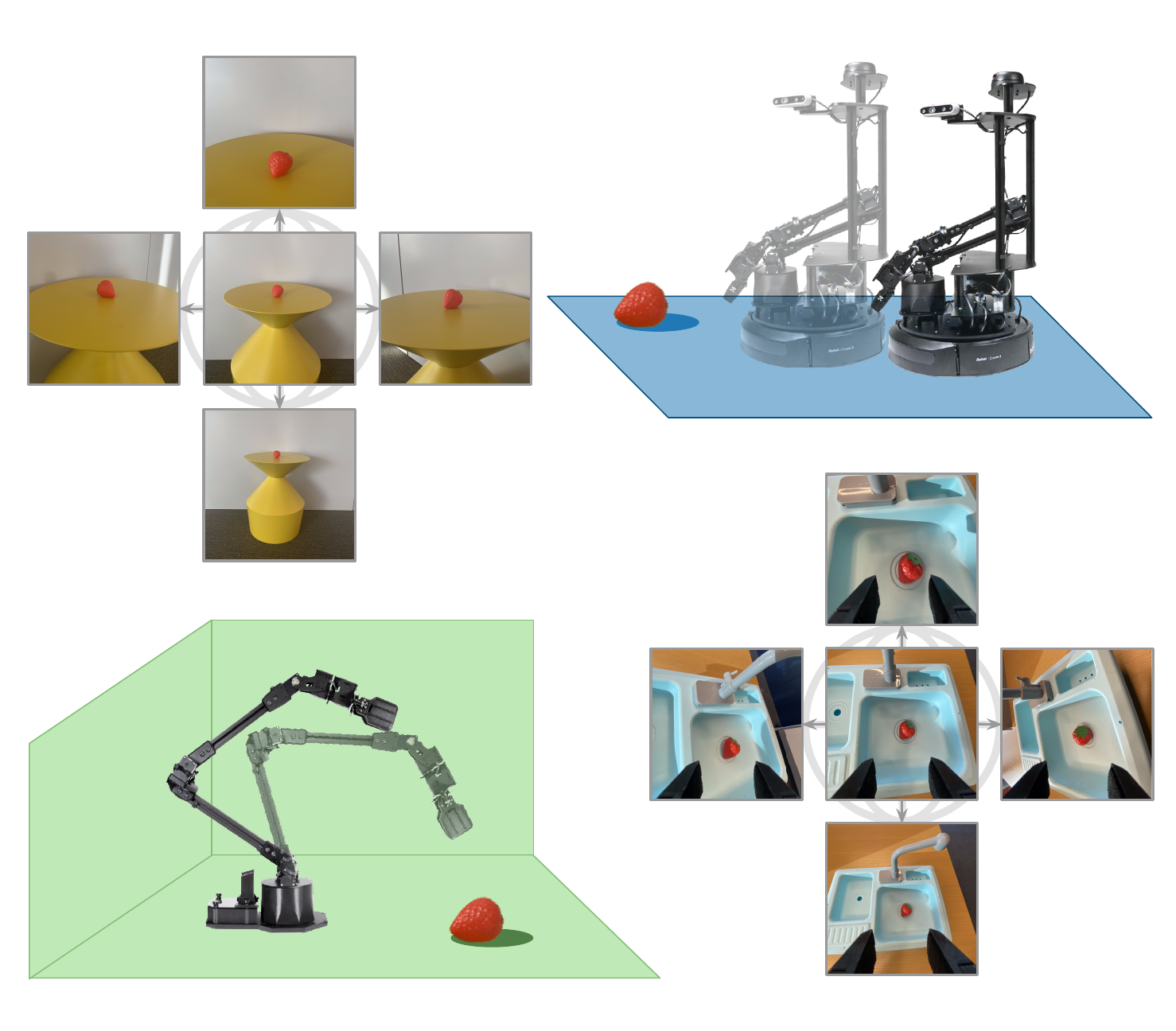}
\caption{\textbf{Unifying Manipulation and Navigation.} Despite having fundamentally different objectives, similar actions lead to similar transformations in the egocentric observations for both manipulators and navigators. We hypothesize that this equivariance can assist training of shared egocentric control policies across both morphologies.
}
\label{fig:equivariance}
\end{figure}

Consider a manipulator reaching for an object from egocentric observations and a navigator trying to reach a waypoint with an onboard front-facing camera. While these tasks span significantly different embodiments and have different action representations, their objectives are similar: to predict a sequence of actions that moves them from the current state to the desired goal. However, their similarities do not end there. Let us define a shared action space coordinate system where $+x$ denotes moving into the image, $+y$ denotes moving left with respect to the image, and $+z$ denotes moving up with respect to the image. Given a certain action and assuming that the robots can move in all directions from any state, the manner in which the observations for a robotic arm transform in response to actions is equivariant to that of a navigation platform. That is, moving ``left" will change the manipulator's observation in the same manner as the navigator's. Figure~\ref{fig:equivariance} depicts this structure, showing that similar actions will lead to similar homographies of a desired object.


Based on this observation, we unify navigation and manipulation into a single task. Consider a trajectory $\tau \in D_{e, z}$. For two observations $o_{i}, o_{j}, \in \tau$ that are temporally close together, define the action $a^{*}$ as the difference in the poses of the cameras that generated these observations. Note that $a^{*}$ is $\textit{agnostic to embodiment}$, meaning that optimizing an action prediction loss $\mathcal{L}(f(o_{i}, o_{j}), a^{*})$, where $f(o_{i}, o_{j})$ tries to prediction $a^{*}$ given its current and goal observation, will not result in fitting different target regardless of whether $o_{i}$ and $o_{j}$ come from a manipulation dataset $D_{e_{m,\cdot}}$ or navigation dataset $D_{e_{n,\cdot}}$. Given that $i$ and $j$ are close enough, we would expect the dataset's action at $o_{i}$, or $a_{i}$ to be similar in direction to $a^{*}$. This is because the robot's motion is continuous, and therefore, locally travels in a straight line. Under these assumptions, training our policy to predict action $a_{i}$ would allow us to learn from $D_{e_{m, 1}} \cup D_{e_{m, 2}} \cup \ldots \cup D_{e_{n, 1}} \cup D_{e_{n, 2}}$ with a single, well-defined objective. 

Of course, in practice, the limitations of physical robotic systems make this equivalence imperfect. First, the dataset action label might not correspond to a straight-line path from the current observation $o_i$ to the goal when the goal is further away -- a manipulator might need to grasp an object, and a ground robot might need to drive around an obstacle.
Secondly, the assumption that any robot can maneuver in any direction from any state is simply untrue. Manipulators are constrained by their joint limits and degrees of freedom. Ground robots cannot move ``upwards" against gravity. In addition, unifying each robot's actions such that moving in a certain direction will correspond to the same change in its egocentric camera parameters is infeasible. The camera parameters of many robotic datasets are not available. Finally, the action magnitudes of different embodiments such as cars and drones may operate on a different scale. However, we still expect that the \emph{local equivariance} provided by this representation should make cross-embodiment training significantly easier.


\subsection{Aligning the Action Coordinate Frames}
To approximate the ideal action space, where the action corresponds to a change in the camera frame, to the greatest extent possible, we learn a \textit{normalized, embodiment-specific} direction. This direction outlines a broad delta-position vector that each robot must move towards, yet allows variations in the scale and strategies that different robots employ to move there. Firstly, for each training dataset $D_{e}$, we normalize the distribution of actions to lie between $-1$ and $1$. This allows the policy to handle different action magnitudes across datasets, which otherwise would cause instability in the action loss. Then, we align the action coordinates across datasets such that each of the action dimensions corresponds to similar transformations in the robots' observations. Ideally, this would be done by first computing the delta Cartesian actions for each robot, then applying a rigid transformation to each action $a$ that maps it to the coordinate frame defined by the camera's extrinsic parameters. However, since most previous large-scale manipulation datasets do not contain this information, in practice, we swap each dataset's action dimensions such that they point in the same direction. Further details are described in the following section.
\begin{figure}
\includegraphics[width=0.5\textwidth]
{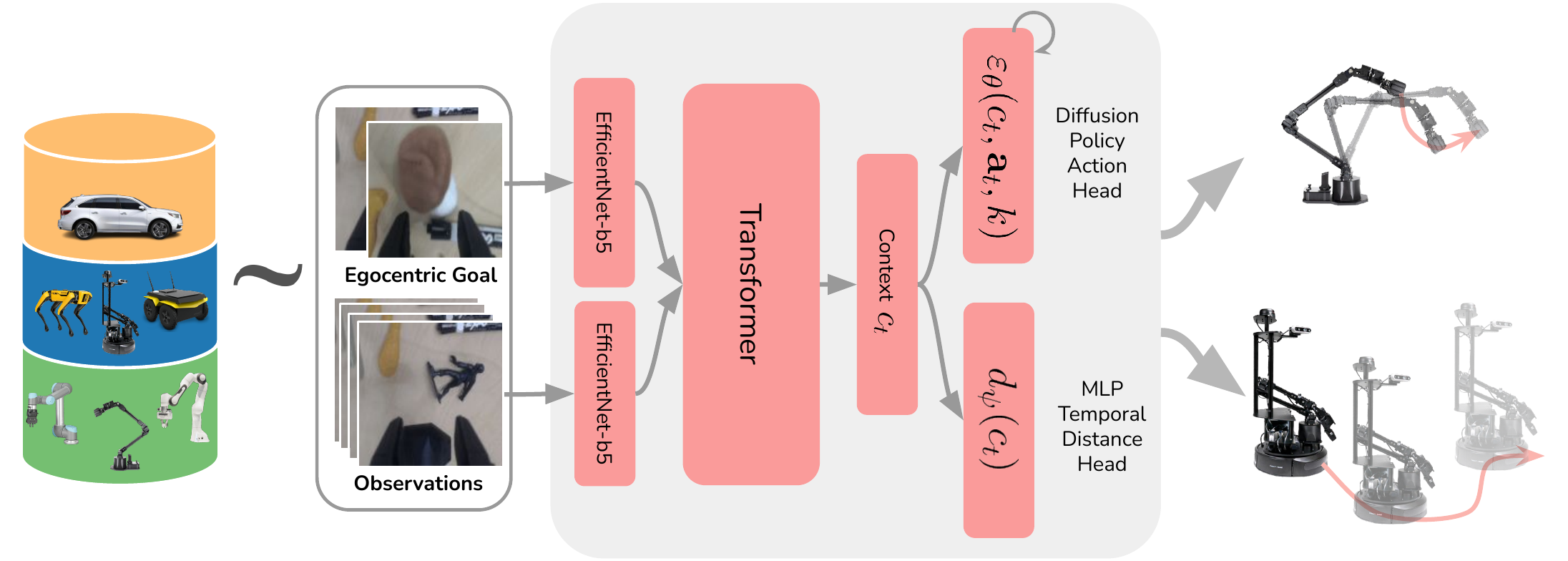}
\caption{\textbf{Policy Architecture.} We use separate observation and goal convolutional encoders to tokenize visual observations, which are passed through a Transformer block. The resulting features are used to predict the temporal distance to the goal $d_\psi$ and future actions $\textbf{a}_t$, using a conditional diffusion process.}
\end{figure}

\subsection{Datasets and Postprocessing}
We train our policy on a mixture of our own small manipulation dataset (see Section~\ref{evaluation:setup}), 9 datasets from OXE \cite{embodimentcollaboration2023open}, a large-scale dataset for wheeled robot navigation, and a large-scale dataset for autonomous driving. The manipulation datasets from OXE include Bridge \cite{ebert2021bridge}, Fractal \cite{brohan2023rt1}, Taco Play \cite{rosete2022tacorl}, Jaco Play \cite{dass2023jacoplay}, Roboturk \cite{mandlekar2019scaling}, NYU Door Opening \cite{VINN}, Viola \cite{zhu2022viola}, Berkeley Autolab UR5 \cite{BerkeleyUR5Website}, and Toto \cite{zhou2023train}. The navigation datasets from GNM \cite{shah2023gnm} include GO Stanford~\cite{hirose2019deep}, SCAND-S/J~\cite{karnan2022socially}, RECON~\cite{shah2021rapid}, Cory Hall~\cite{kahn2018gcg}, Seattle~\cite{shaban2021semantic}, and TartanDrive~\cite{triest2022tartan}. Additionally, we also train on SACSoN~\cite{hirose2023sacson} and Berkeley Deep Drive~\cite{yu2020bdd100k}. Each of these datasets is converted to the RLDS format \cite{ramos2021rlds}. We upweight the frequency at which navigation data appears such that $50\%$ of the entire data mixture is navigation data and $50\%$ is manipulation. This ensures that the policy fits evenly to the two domains, which is important not to degrade the performance of one domain in favor of another. For the manipulation share of data mixture, we weight the datasets by a similar split to prior work \cite{octo_2023}. For navigation, we weight the datasets by their relative number of trajectories. For further information, we refer the reader to  \cref{appendix:dataprocessing}.

We post-process the manipulation datasets by aligning the coordinate frames of the actions as described in the previous section. Note that OXE does not contain consistent action coordinate systems for each robot. After converting each of the datasets to delta Cartesian actions, the dimension $0$ can correspond to the robot moving left, right, or forward depending on their control scheme. Therefore, we alter the datasets inside our mixture by correcting each of the dimensions of the action coordinates frames such that each dimension of the action corresponds to the same general direction of the end effector. This is done by manually sampling (observation, action, next observation) pairs for each dataset and observing the change in robot pose with respect to the actions. For coordinate systems that don't align with the coordinate system of our manipulation dataset, we swap the dimensions and signs of the actions to be more consistent.
For manipulation, we use a $7$-dimensional action space with zero-indexed dimensions $0-2$ as delta Cartesian actions, $3-5$ as delta rotations, and $6$ as gripper open/close. 

Each of the navigation datasets contains a sequence of egocentric images and states containing a position $p = (x, y)$ as well as a yaw $\phi$. For each action, we subtract the current state from the next $5$ future states, then transform these differences with a rotation matrix defined by the yaw in order to get egocentric waypoints. These actions are transformed into the manipulation coordinate frame as described in the previous section. Since the egocentric camera is pointed downwards and towards the gripper for our manipulation tasks, we map the ``forward" $+y$ axis in our navigation datasets to the $-z$ downwards direction in our manipulation datasets. We also map the ``left" $+x$ direction in our navigation dataset to the ``left" $+y$ direction in our manipulation datasets. Therefore, we translate navigation action $a \in D_{e_{n}, t}$ to $(0, a[1], -a[0], 0, 0, 0, 0)$.

\subsection{Policy Architecture}

To fit a single policy to datasets across a wide variety of embodiments, the neural architecture must be both scalable and expressive. We design our model to scale up a simple transformer backbone. At a high level, we want our model to process its observations using some encoder, feed its embeddings into a transformer, and then output both an action and the distance to its goal. We use combined insights from previous large-scale robot models in manipulation \cite{brohan2023rt1} and navigation \cite{shah2023gnm} to motivate our design decisions. Firstly, our choice of observation encoders is EfficientNet ConvNets \cite{tan2020efficientnet}, which have been used successfully in robot learning for both navigation and manipulation. For our action output head, we chose to use a diffusion policy \cite{chi2023diffusion} to account for noise in human demonstrations as well as different strategies that may exist to reach the goal state. In addition, we both incorporate history and predict future actions, parameterizing our policy as $\pi(a_{t: t + k - 1 } | o_{t - c + 1: t}, o_{g})$. This decision stems from prior work indicating that policies trained with past states and future actions exhibit significant improvements in their ability to fit to teleoperated demonstration data \cite{robomimic2021, zhao2023learning}.

Our heterogeneous cross-embodiment model consists of five different components: two observation encoders, a transformer, a diffusion policy action head \cite{chi2023diffusion}, and an MLP distance prediction head for navigation with topological graphs. Similar to the scheme proposed by NoMaD \cite{sridhar2023nomad}, we encode the observation history $o_{t-k:t}$ with an EfficientNet-b5 encoder. We then concatenate the current observation $o$ and goal observation $o_{g}$ with a separate EfficientNet-b5 encoder in a channel-wise manner. The resulting embeddings are then concatenated and fed to the transformer to get an action and distance prediction. The action prediction is reshaped into a tensor of size $(b, n, 7)$, while the distance prediction is reshaped into a tensor of size $(b, 1)$, where $b$ is the batch size and $n$ is the number of actions we predict into the future. Note that this distance prediction is used by navigation policies to localize the robot with respect to a topological map, but not used by manipulation policies. Regardless, distance to the goal is a well-defined task, even in manipulation.

We train our policy with diffusion denoising loss
\begin{center}
$\mathcal{L}_{\text{diffusion}}(\theta,  \psi) = ||\epsilon_{k} - \epsilon_{\phi}(f_{\theta}(o_{t-k:t}, o_{g}), a_{t}^{0} + \epsilon_{k}, k)||_{2}^{2},$ 
\end{center}
and a distance prediction loss
\begin{center}
$\mathcal{L}_{\text{distance}}(\theta, \psi) =  ||d_{\psi}(f_{\theta}(o_{t-k:t}, o_{g})) - d_{t}||_{2}^{2}.$
\end{center}
Our overall objective is the weighted combination of these two losses:
\begin{center}
$\mathcal{L}(\theta, \phi, \psi) = \mathcal{L}_{\text{diffusion}}(\theta,  \psi) + \lambda \mathcal{L}_{\text{distance}}(\theta, \psi).$
\end{center}
In practice, we find that $\lambda = 0.001$ is a reasonable value, which ensures that the distance head is trained but doesn't interfere with the action loss. $f_{\theta}(o_{t-k:t}, o_{g})$ denotes the observation encoder and transformer, $\epsilon_{\phi}$ denotes the noise prediction head, and $d_{\psi}$ denotes the noise prediction head. The noise prediction network tries to predict the noise at iteration $k$, or $\epsilon_{k}$ from the noisy action $a_{t}^{0} +  \epsilon_{k}$, where $a_{t}^{0}$ denotes a flattened action from the dataset. In addition, $d_{t}$ denotes the distance in timesteps from the current observation and goal observation. The goal image is sampled uniformly at random $20$ to $40$ timesteps into the future from the current observation. This firstly provides local goals such that the direction between the current observation and the goal and be ascertained. Secondly, this allows our method to scale to datasets within OXE that contain long sequences of observations.

Since the majority of OXE consists of third-person observations without wrist-image counterparts, when sampling new batches, we select between these two viewpoints uniformly at random if both exist or use the available one. Our experiments show that in certain domains, co-training with 3rd-person images can greatly increase the success rate of the policy. For these results, we refer the reader to Appendix~\ref{appendix:additional}. As more large-scale wrist-camera datasets are released in the future, we believe that this gap will close.

%% file: text/5-evaluation.tex
\begin{figure*}
    \centering
    \includegraphics[width=0.92\textwidth]
    {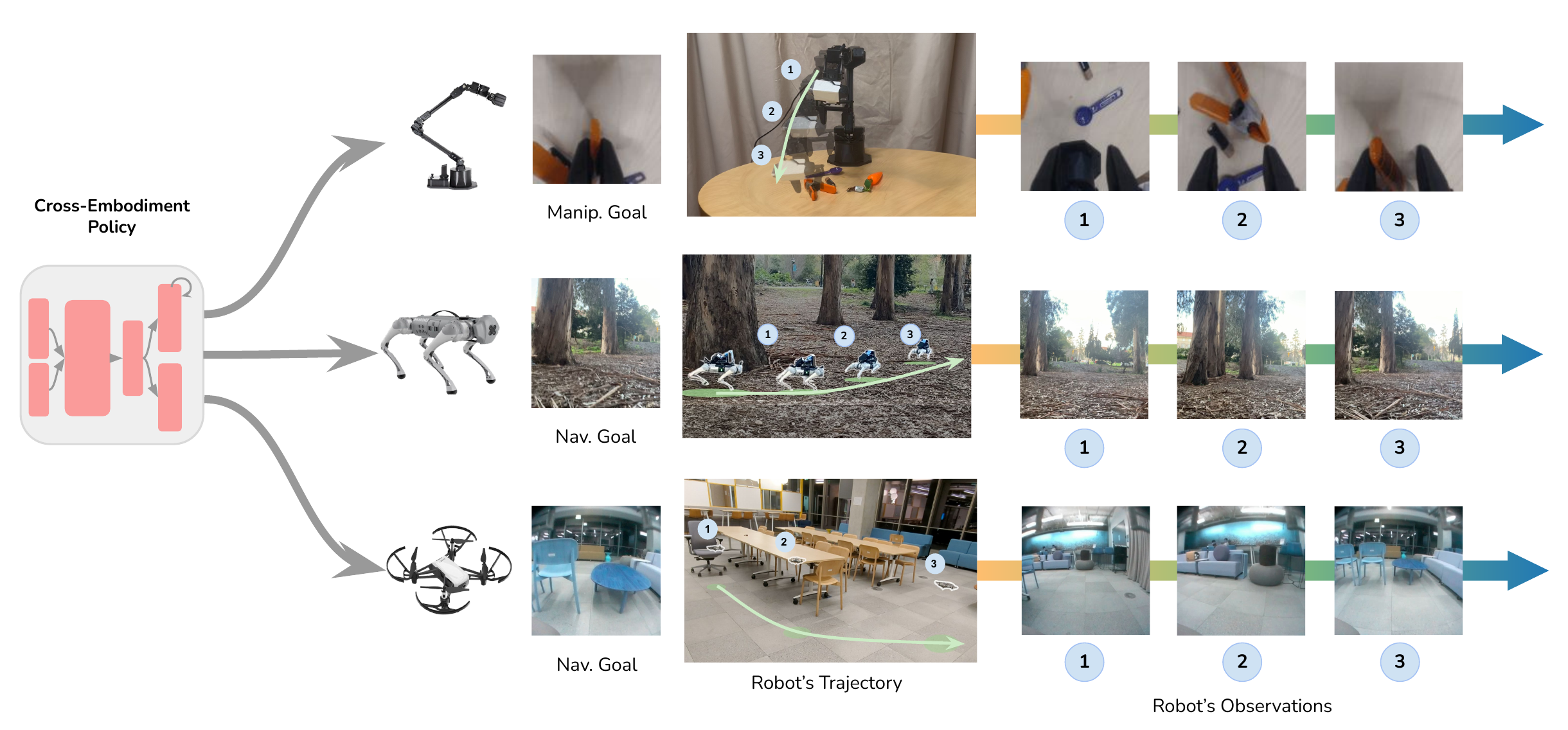} 
    \caption{Qualitative examples of the \emph{same policy checkpoint} deployed on a tabletop manipulator solving the ``Cluttered Grasp'' task (top), a quadruped navigating to a goal in a forest (middle), and a drone navigating a cluttered office environment (bottom).}
\end{figure*}

\section{Evaluation}

Our goal is to evaluate the performance of heterogeneous cross-embodiment policies in solving real-world manipulation and navigation tasks on a variety of embodiments. In addition, we aim to investigate the possibility of knowledge transfer across these embodiments. To this end, we seek to answer the following questions: 
\begin{enumerate}
    \item Can a single goal-conditioned policy successfully control widely varying embodiments for both navigation and manipulation? 
    \item Can co-training with navigation data provide generalization benefits to manipulation policies? 
    \item How does navigation data help manipulators generalize? 
    \item What kind of navigation data enables better transfer to manipulation tasks? 
    \item Can co-training with manipulation data provide generalization benefits to navigation policies?
    \item Can heterogeneous cross-embodiment policies generalize zero-shot to new embodiments? 
\end{enumerate}


\subsection{Evaluation Embodiments}
In order to demonstrate the ability of our policy to fit to a wide range of embodiments, we evaluate on five low-cost, open-source robot manipulators and mobile robots, including a mobile manipulator (see~\cref{fig:teaser}):
\begin{itemize}
    \item \textbf{WidowX250S:} A 6-DoF robotic arm with a parallel jaw gripper, with a camera on the wrist.
    \item \textbf{LoCoBot:} An indoor mobile robot with a forward-facing camera.
    \item \textbf{Clearpath Jackal:} A fast mobile robot with a forward-facing camera capable of moving indoors and outdoors.
    \item \textbf{DJI Tello:} A quadrotor with a forward-facing camera.  
    \item \textbf{Unitree Go1:} A quadruped with a forward-facing camera. 
    \item \textbf{Mobile ALOHA:} A bimanual mobile manipulation platform with a mobile base and two ViperX arms~\cite{fu2024mobile}. Each arm has a camera on the wrist, and the base has a forward-facing camera.
\end{itemize}

\subsection{Manipulation and Navigation Tasks}
We aim to test whether navigation data can provide generalization benefits to manipulators. We evaluate our method on $5$ tasks outlined below. These tasks were constructed to require information from the goal image for the policy to obtain high success.
\begin{enumerate}
    \item \textbf{Two-object Reaching.} A simple environment with two different objects to the left and to the right of the manipulator. Given the goal image, the manipulator needs to move to the correct object, similar to the setup in navigation.
    \item \textbf{Cluttered Grasp.} A grasping task where the robot has to pick the correct object out of $5$ different objects seen in its training data. The positions and interactions between the objects are randomized and may not be seen in the training dataset. 
    \item \textbf{Novel Cluttered Grasp.} A cluttered grasping task with $5$ held-out objects. The manipulator needs to pick the correct object as specified by the goal image.
    \item \textbf{Toy Kitchen.} A more semantically meaningful environment where the robot needs to pick up a strawberry and eggplant from the sink. This exact environment is not explicitly seen in the training data, although similar toy kitchen setups exist in the BRIDGE dataset.
    \item \textbf{Shelf Manipulation.} A task where the WidowX250S has to pick out the right object from a shelf compartment. This location of the shelf is randomized. This task evaluates the ability of the policy to generalize to variations in distance to an axis perpendicular to the egocentric camera while avoiding colliding with the shelf.  
\end{enumerate}

To evaluate our navigation policies, we chose two novel locations that were not seen in any of the training datasets.
\begin{enumerate}
    \item \textbf{\BAIR Hallway:} A navigation task in a hallway with clutter. The robot must navigate around two corners without colliding with obstacles and stop at the last goal location.
    \item \textbf{\BAIR Kitchen:} A navigation task in a more open kitchen environment. As with the previous task, the robot must navigate to the final goal image location without colliding with obstacles. 
\end{enumerate}

Before rolling out our policy for manipulation tasks, we collect a goal image for each task by rearranging the environment and teleoperating the manipulator to the desired state. For navigation, we create a topological map $\mathcal{M}$ by recording the robot's observations with a frequency of $4$ Hz while moving the robot base throughout the environment. We ensure that this map has sufficient coverage of the locations which the robot may traverse during evaluation time. 

\begin{figure*}
    \includegraphics[width=\textwidth]
    {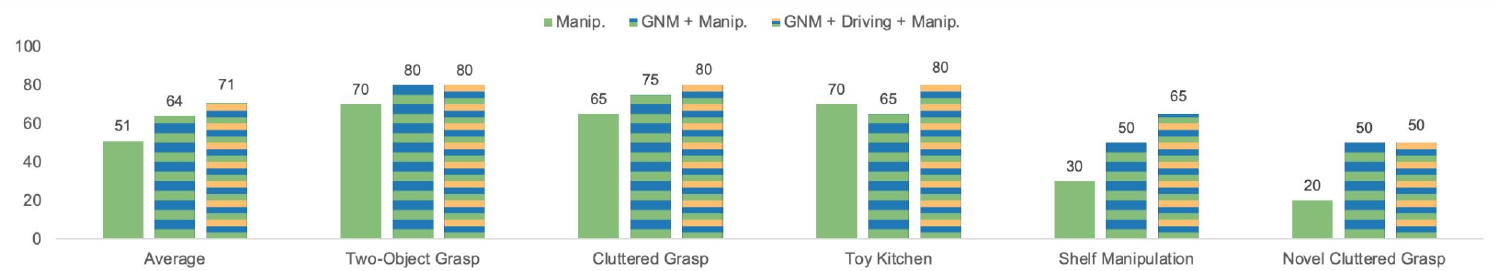} 
    \caption{\textbf{Does navigation help manipulation?} By aligning action coordinate frames, training on navigation and driving datasets results in a $20\%$ improvement across five challenging tabletop manipulation tasks (success \% on y-axis).}
    \label{fig:manipeval}
\end{figure*}

\subsection{Experiment Setup}
\label{evaluation:setup}
To investigate the impact of certain navigation datasets on manipulation and vice versa, we ablate including the various datasets. For clarity, we label the GNM dataset as \textbf{GNM}, the BDD100k dataset as \textbf{Driving} and the combination of OXE and our own tabletop manipulation dataset as \textbf{Manip}. Although we can obtain successful results on our navigation robots by solely using prior open-source datasets, collecting our own dataset specific to our embodiments and control schemes was necessary for manipulation. While navigation policies generalize in zero-shot to new embodiments, current manipulation policies require in-distribution data for the target embodiment and control scheme in order to identify the correct action space and understand important visual features of the robot ~\cite{embodimentcollaboration2023open}. Therefore, to ensure a nonzero success rate for our method on our embodiments, we collect our own manipulation data. This dataset contains expert demonstrations on the WidowX250s and ViperX300s collected via VR teleoperation. The WidowX250s dataset consists of manipulation with a set of $5$ training objects. For each in-distribution object, we collect $50$ grasping demonstrations while randomizing the locations of the other objects. The ViperX300s dataset consists of $50$ pick and place trajectories each for a set of $3$ training objects. This entire dataset spans $400$ trajectories collected over the course of $8$ hours.

\section{Analysis}

\subsection{Can a single goal-conditioned policy successfully control widely varying embodiments for both navigation and manipulation? }
\begin{figure*}
    \includegraphics[width=\textwidth]
    {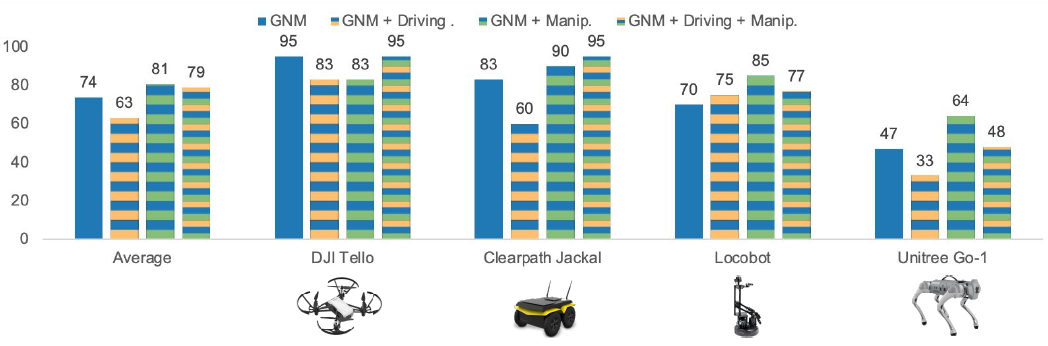} 
    \caption{\textbf{Does manipulation help navigation?} Across three different robots in challenging indoor and outdoor environments, adding manipulation datasets leads to $5-7\%$ improvement in navigation performance (success \% on y-axis).\label{fig:nav}}
\end{figure*}

We report evaluation results for our method for manipulation in Figure~\ref{fig:manipeval} and for navigation results in Figure \ref{fig:nav}. Our method obtains an average of $71\%$ success rate over $5$ different manipulation tasks, and an average of $80\%$ success rate on $2$ navigation tasks each on $4$ different embodiments. This demonstrates our policy's ability to fit to both manipulation and navigation datasets. In addition, our policy can identify and output an action for the appropriate embodiment given its current and goal observations. For example, when given observations from a mobile manipulator, the policy outputs a waypoint in the second and third dimensions along with values that are near $0$ in the other dimensions. This is consistent with the scheme we used to align manipulation and navigation actions. 


\subsection{Can co-training with navigation data provide generalization benefits to manipulation policies?}
Figure \ref{fig:manipeval} shows the success rate of various dataset mixtures on manipulation tasks. Training our policy on a manipulation and navigation data split had a $20\%$ greater success rate over $5$ tasks compared to training only on manipulation data. The largest gap in performance between the joint navigation-manipulation policies and the manipulation-only policies was in the Novel Cluttered Grasp and Shelf Manipulation scenarios. These scenarios involve spatial reasoning in novel environments (e.g., in Shelf Manipulation, the policy must learn which actions don't collide with the shelf), requiring the policy to understand the location of its current state with respect to the goal. 

For the Cluttered Grasp tasks, the gap in performance between the joint navigation-manipulation policy is larger in the out-of-distribution variant than the in-distribution variant. A plausible explanation could be that the navigation data regularizes the policy's intermediary representations to capture relative spatial information between current and goal images. In Shelf Manipulation, the robot needs to grasp an object located on a shelf with randomized positions. This requires the robot to avoid colliding with the shelf as well as gauge its distance to the object, which is fundamentally similar to the collision avoidance task in ground navigation. Gauging object distance is analogous to testing the robustness to a change in table height in tabletop manipulation, which previous works have identified as a common distribution shift artifact leading to failure~\cite{hsu2022visionbased, xie2023decomposing}.

\subsection{Can co-training with manipulation data provide generalization benefits to navigation policies?}
Figure~\ref{fig:nav} reports our navigation results. Each dataset mixture was evaluated on four different robots across two indoor domains, then averaged to get a success rate. Three of these robots--the LoCoBot, Jackal, and Unitree Go1--were present in the training dataset, while the DJI Tello is a novel embodiment. Due to a difference in the camera lens used by the DJI tello, we noticed that the performance of the drone degraded significantly in environments without clear markers of corners. Therefore, while the LoCoBot, Clearpath Jackal, and Unitree Go1 were evaluated in everyday environments, the DJI Tello had to be evaluated in scenes with bright objects to indicate corners at which to turn. 

On the Jackal, LoCoBot, and Unitree Go1, we observed greater success rates for policies that were co-trained with navigation and manipulation data, with \textit{GNM + Driving + Manip} being  $12\%$, $7\%$ and $1\%$ better than \textit{GNM-only} respectively. With \textit{GNM + Manip}, the success rate is $7\%$, $15\%$, and $17\%$ respectively. On the DJI Tello, the \textit{GNM + Driving + Manip} performs similarly to the \textit{GNM-only} policy, each with $95\%$ success rate. Averaged over the embodiments, the policy trained with manipulation data had $5-7\%$ higher success than the navigation-only policy. While we qualitatively observed that these policies had better estimates for the closest node and had less collision with the environment, we acknowledge that the $5-7\%$ difference is not particularly large and can potentially be explained due to the variance between evaluation runs. We believe that this gap will widen more in-the-wild manipulation data or mobile manipulation data in the future.
 
\subsection{How does navigation data help manipulators generalize?}
\begin{table}[!htb] 
    \centering
    \begin{tabular}{lccc}
    \toprule
    \textbf{Datasets} & \textbf{Cluttered} & \textbf{Novel Cluttered} & \textbf{Shelf} \\
    \midrule
    GNM + BDD + Manip & 0.749& 0.740& 0.644\\
    GNM + Manip & 0.743& 0.735& 0.648\\    
    Manip-only & 0.733& 0.703& 0.614\\
    \bottomrule
    \end{tabular}
     \caption{\textbf{Embedding Analysis.} Transformer features for policies co-trained with all datasets have a stronger linear correlation ($R^2$ coeff.) with the temporal distance than manip.-only policies.}
    \label{table:cca}
\end{table}

To investigate our hypothesis that navigation data can help a manipulator understand its position with respect to its goal, we collected a small dataset of trajectories for each policy and computed our policy's embedding before the action and distance heads. Then, we ran canonical correlation analysis between these features and the temporal distance between the current state and the goal and recorded the resulting coefficients of determination ($R^{2}$) in Table~\ref{table:cca}. Our results show a positive correlation between ratio of the coefficient of determination between data splits and the ratio of the success rates on manipulation tasks. In particular, the ratio of the performance of the full manipulation-navigation policy and the manipulation-only policy are $1.230$, $2.75$, and $1.14$ on the Cluttered Grasp, Novel Cluttered Grasp, and Shelf Manipulation tasks. The ratio of the $R^{2}$ coefficients between these policies are $1.011$, $1.061$, and $1.049$ respectively. The observation that higher correlation values are indicative of better performance supports our hypothesis that goal-conditioned policies co-trained with navigation data better understand their relationship with a goal image.

To further examine whether information from the goal image is essential to transferring navigation data to manipulation, we ran an ablation of our method without goal-conditioning. Table \ref{table:nogoalconditioning} shows the results of these policies in the novel cluttered grasp task. Note that while goal-conditioned experiments record the proportion of trials in which the robot grasped the $\textit{correct}$ object, the unconditioned experiments record the proportion of trials in which the robot grasped \textit{any} object. The gap in performance between adding \textit{GNM + Manip} and \textit{Manip-only} in the goal-conditioned and the unconditioned policies are $35\%$ and $5\%$ respectively. Operating under the assumption that the diffusion policy is powerful enough to model the different possible tasks from the current observation without conditioning on the goal image, we can conclude that information transfer between manipulation and navigation policies is negligible without goal conditioning. 

\subsection{What kind of navigation data enables better transfer to manipulation tasks? }
We ablate which datasets inside of GNM \cite{shah2023gnm} we co-train with to investigate which types of navigation environments are more conducive to transfer to manipulation. We provide further information about the location and length of each dataset in Appendix \ref{appendix:gnm}. Figure~\ref{fig:nav_data_ablate} indicates that trail-like outdoor datasets such as Tartan and Seattle do not provide positive transfer to manipulation scenarios. Meanwhile co-training with indoor manipulation environments such SACSoN and GO Stanford lead to significant improvements in manipulation performance. We hypothesize that this difference is due to the presence of sharper angles and objects with well-defined boundaries in indoor navigation data.

\begin{table}[!htb]
    \centering
    \begin{tabular}{cccc}
    \toprule
    \textbf{GNM + Manip} & \textbf{Manip-only} & \textbf{GNM + Manip} & \textbf{Manip-only} \\
    \textbf{GC} & \textbf{GC} & \textbf{UC} & \textbf{UC} \\
    \midrule
    $55\%$ & $20\%$ &  $45\%$ & $40\%$\\
    \bottomrule
    \end{tabular}
     \caption{\textbf{Is goal-conditioning important for transfer?} There is a $30\%$ higher gap in performance between goal-conditioned (GC) co-trained policies and manipulation-only policies compared to unconditioned (UC).}
    \label{table:nogoalconditioning}
\end{table}

\begin{figure*}
    \includegraphics[width=\textwidth]
    {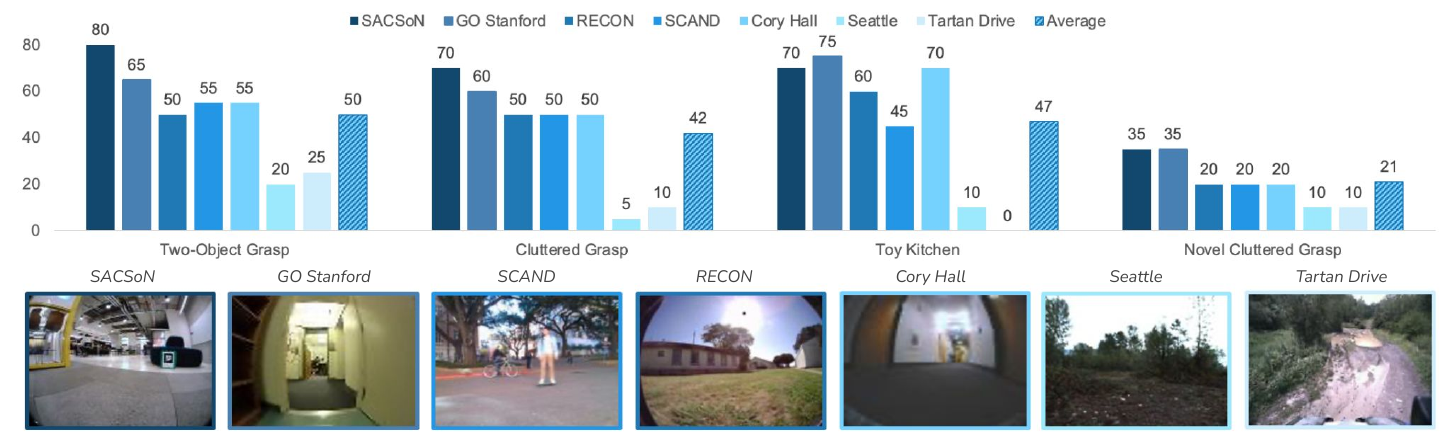} 
    \caption{\textbf{What type of navigation data helps positive transfer?} Manipulation policies co-trained with indoor and outdoor navigation data on sidewalks perform better than policies co-trained on forest trails and off-road environments.}
    \label{fig:nav_data_ablate}
\end{figure*}
\subsection{Can heterogeneous cross-embodiment policies generalize zero-shot to new embodiments?}



By training our policies with both manipulation and navigation data, heterogeneous cross-embodiment policies can allow robots that require both manipulation and navigation to leverage preexisting domain-specific large-scale datasets. To test the limit of this generalization capability, we evaluate our policy on the Mobile Aloha platform \cite{fu2024mobile}. While the robot is capable of bimanual mobile manipulation, we simplify the platform by only using the right manipulator. To obtain actions for both navigation and manipulation, we run our policy twice: one from an egocentric camera mounted to the robot arm, and one from the navigation camera. We threshold the magnitude of the policy's action prediction to determine when to run the manipulation policy. Namely, if the policy is close to the final goal image and the actions are small, we allow the manipulation policy to control the robot arm.

\begin{figure}
    \centering
    \includegraphics[width=0.92\columnwidth]{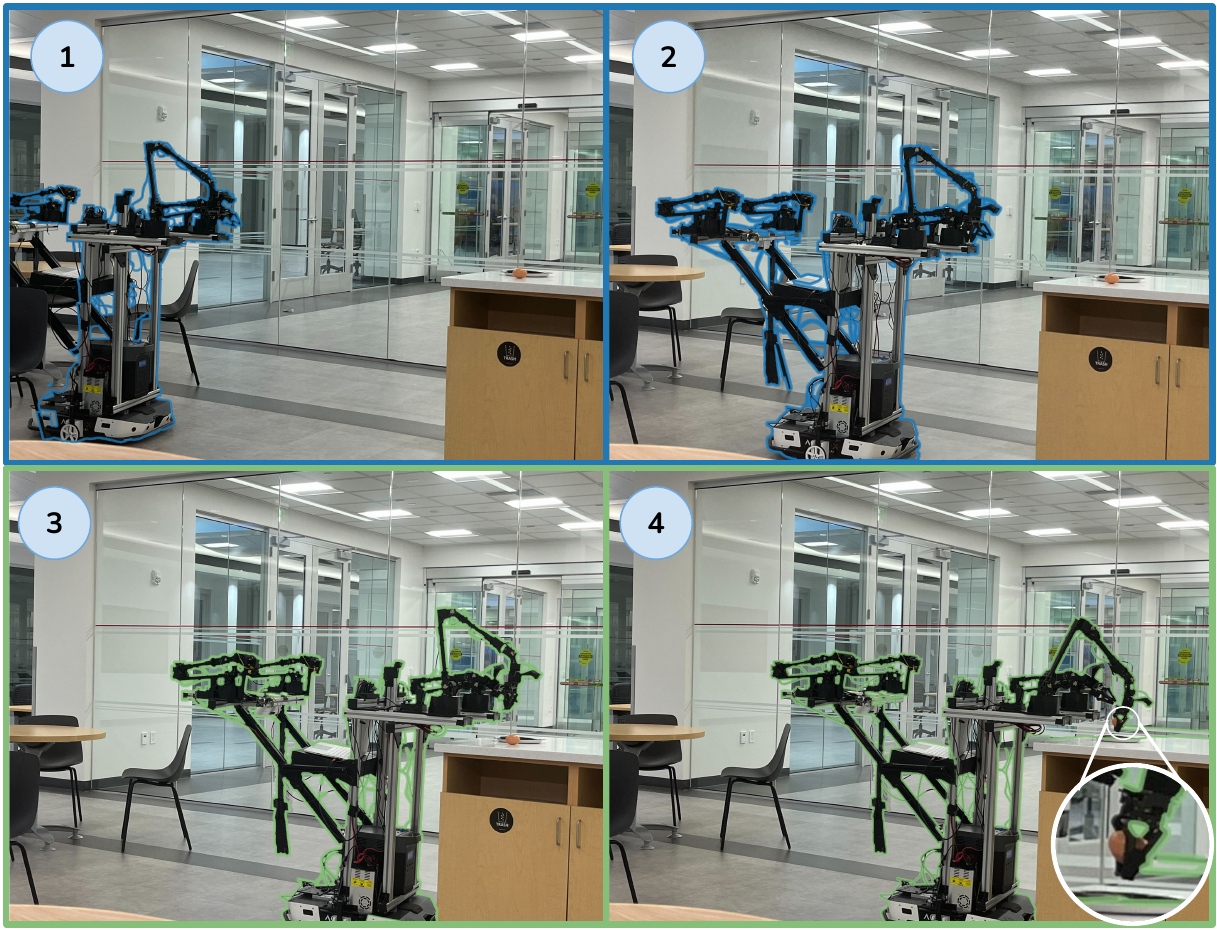}
    \caption{Our cross-embodiment policy trained on manipulation and navigation data zero-shot generalizes to a mobile manipulator, succeeding in the ``Egg Nav/Pick/Place'' task.}
    \label{fig:mobile_aloha}
\end{figure}

We evaluate our policy on the $\textit{Egg Nav/Pick/Place}$ task, where the robot has to approach a table, pick up an egg, and place it onto a plate (see Fig.~\ref{fig:mobile_aloha}). Despite the fact that neither the table nor the egg was seen in the training data of the policy, the robot achieves a $50\%$ success rate, demonstrating the method's effectiveness in controlling a new embodiment. Qualitatively, the robot succeeds when the robot manipulator is located in a good position with respect to the object. However, small changes in the mobile base can elicit large changes in position of the robot arm with respect to the scene, and the robot can fail to move its base such that its arm into a favorable position before executing the task.

\begin{table}[!htb]
    \centering
    \begin{tabular}{cc}
    \toprule
    \textbf{Datasets} \textbf{Egg Nav/Pick/Place} \\
    \midrule
    GNM + Driving + Manip  & 50\% \\
    \bottomrule
    \end{tabular}
     \caption{\textbf{Zero-shot Embodiment Generalization Experiments.} Our policy demonstrates the ability to transfer to new embodiments that were not seen in the training data.}
    \label{table:zeroshot}
    \vspace*{-1em}
\end{table}

%% file: text/7-appendix.tex
\clearpage
\newpage

\section{Appendix}
\subsection{Data Postprocessing}
\label{appendix:dataprocessing}

\subsubsection{Dataset Mixture}
The following table shows the data mixture we used to train our policy: \\

\begin{table}[!htb]
    \centering
    \begin{tabular}{c|c}
    \toprule
    \textbf{Datasets} & \textbf{Split} \\
    \bottomrule
    Ours & $1.3\%$ \\
    Bridge \cite{ebert2021bridge}& $5.3\%$ \\
    Fractal \cite{brohan2023rt1} & $1.3\%$ \\
    Taco Play \cite{rosete2022tacorl}& $2.6\%$ \\
    Jaco Play \cite{dass2023jacoplay} & $5.3\%$ \\
    Roboturk \cite{mandlekar2018roboturk} & $2.6\%$ \\
    NYU Door Opening \cite{VINN} & $1.3\%$ \\
    Viola \cite{zhu2022viola} & $13.3\%$ \\
    Berkeley Autolab UR5 \cite{BerkeleyUR5Website} & $5.3\%$ \\
    Toto \cite{zhou2023train} & $5.3\%$ \\
    GNM \cite{shah2023gnm} & $46.3\%$ \\
    BDD100k \cite{yu2020bdd100k}& $10\%$ \\
    \bottomrule
    \end{tabular}
     \caption{\textbf{Data Splits} }
    \label{table:manipeval}
\end{table}
Ours denotes a manipulation dataset we collected via teleoperation containing $300$ WidowX and ViperX trajectories. We weight the datasets such that navigation accounts for roughly half of the data split. This ensures that the policy would be able to fit our control scheme and egocentric viewpoint. Since BDD100k is comparatively larger than the rest of the datasets, we subsample this dataset to make data loading more efficient. 

\subsubsection{Data Loading}
To load our datasets in an efficient manner, we use a tensorflow dataloader with the RLDS format \cite{ramos2021rlds}. Although all of RT-X is already in this format, we write conversion scripts for the navigation datasets as well. We filter each of the datasets for the relevant observations and actions before loading them. Since RLDS shards entire trajectories together, this is significantly more efficient than filtering the datasets after loading. To determine the correct coordinate frame alignment, we visualize trajectories from each dataset and analyze how transformations in observations correspond to changes in each dimension of the actions. Then, after loading each dataset, we map each action to its dataset-specific coordinate frame transformation.

\subsection{GNM Navigation Data}
The following table describes the datasets inside GNM that we used to train our policy, as well as their environment and split inside the dataset mixture. Further information can be found in the GNM paper \cite{shah2023gnm}. Each of these datasets its weighted proportionally to its size.
\label{appendix:gnm}
\begin{table}[!htb]
    \centering
    \begin{tabular}{c|c|c}
    \toprule
    \textbf{Datasets} & \textbf{Environment}  & \textbf{Split} \\
    \bottomrule
    SACSoN \cite{hirose2023sacson} & Indoors & $7\%$\\
    GO Stanford \cite{hirose2023sacson} & Indoors & $9.8\%$\\
    SCAND \cite{karnan2022socially} & Outdoors, sidewalk & $6.3\%$ \\
    RECON \cite{shah2021rapid} & Outdoors, off-road & $17.6\%$ \\
    Cory Hall \cite{kahn2018gcg}& Indoors & $1.4\%$\\
    Seattle \cite{shaban2021semantic}&  Outdoors, trail & $0.7\%$\\
    Tartan Drive \cite{triest2022tartan} &  Outdoors, trail & $3.5\%$\\
    \bottomrule
    \end{tabular}
     \caption{\textbf{GNM Splits} }
    \label{table:manipeval}
\end{table}

\subsection{Navigation Details}
\label{appendix:gnm}
In this section, we provide further details on how we evaluate our policy on a mobile robot. Firstly, we move the robot around the environment and record a video at a $4hz$ frequency. The video is then converted into a topological map of goals. This map contains images that the navigation policy may potentially reach when navigating in the environment. During evaluation time, the agent first estimates the temporal distance of all the images in the topological map with respect to its current location. The robot then chooses a goal that is close to its current state, but also progresses towards its final destination. This is done by finding the image that minimizes the temporal distance prediction, and then choosing a goal image that is closer to its destination, but not too far. Finally, the robot uses the output of the policy as a waypoint. The waypoint is converted into linear and angular velocities, and then passed as commands to the robot's servos.

\subsection{Hyperparameters}
We train our policy with the following hyperparameters on a single 48G gpu for around 48 gpu-hours. 
\begin{table}[!htb]
    \centering
    \begin{tabular}{c|c}
    \toprule
    \textbf{Hyperparameter} & \textbf{Value} \\
    \bottomrule
    Batch Size & 256 \\
    Learning Rate & 1e-4 \\
    Learning Rate Scheduler & Cosine \\
    Observation Encoder & EfficientNet-b5 \\
    Transformer Attention Heads& 8 \\
    Transformer Attention Layers & 8 \\
    Transformer Hidden Dim& 2048 \\
    Diffusion Iterations& 10 \\
    Diffusion UNet Dims& 128, 256, 512, 1024 \\
    \bottomrule
    \end{tabular}
     \caption{\textbf{Hyperparameters} }
    \label{table:manipeval}
\end{table}

\begin{figure*}[h]
\includegraphics[width=\textwidth]{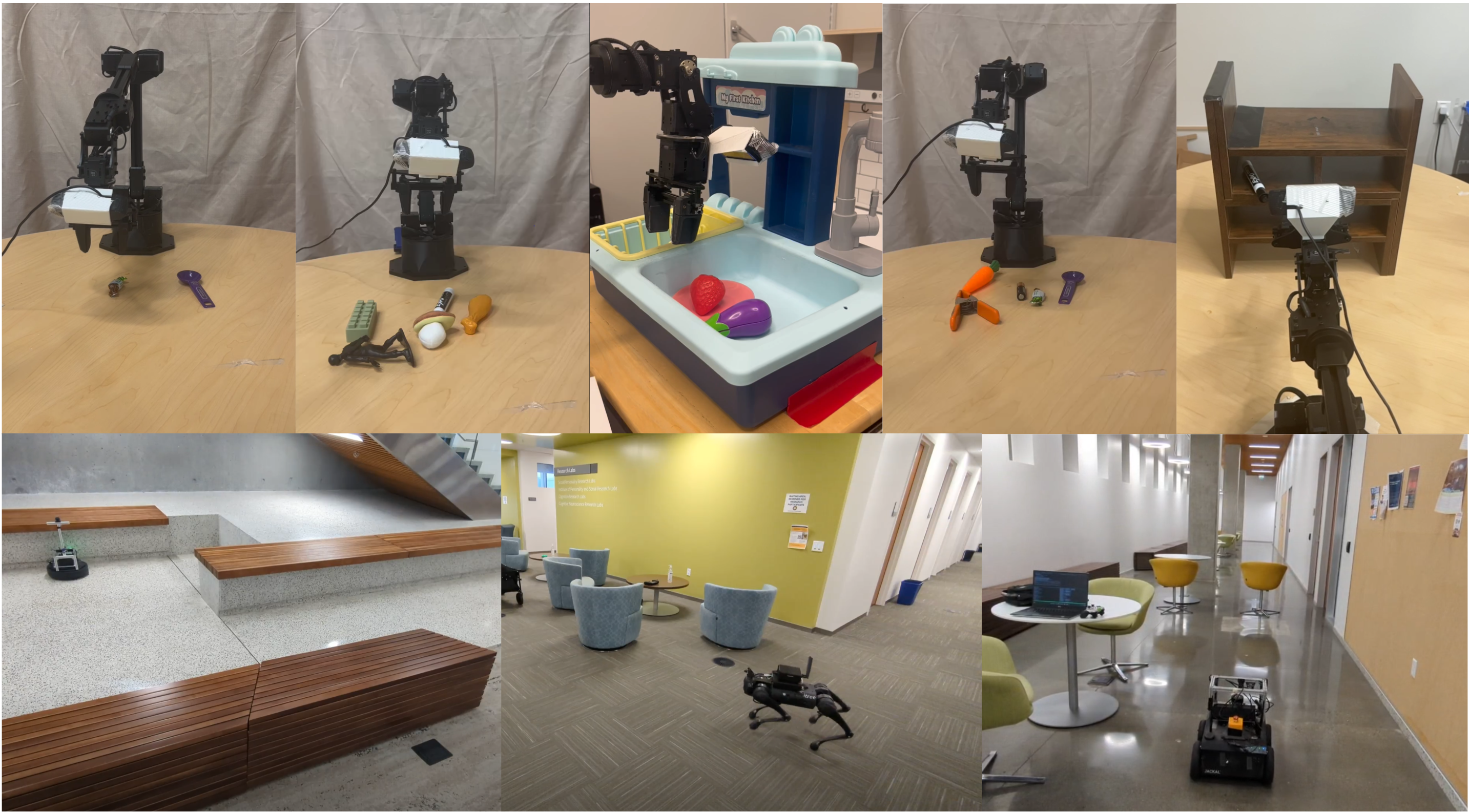}
\caption{\textbf{Navigation and Manipulation Tasks. } We evaluate our policy's manipulation performance on the two-object grasp, cluttered grasp, toy kitchen, cluttered grasp novel, and shelf manipulation tasks (from left to right). To evaluate our policy's navigation performance, we run our policies on the LoCoBot, Unitree Go-1 and ClearPath Jackal (from left to right). }
\label{fig:navmaniptasks}
\end{figure*}

\subsection{Egocentric Cotraining Ablation}
\label{appendix:additional}
We compare the performance of our method with and without co-training with third person viewpoints. Our results show a drop in performance when training with only wrist observations. This gap is more pronounced in the Toy Kitchen and Cluttered Grasp Novel settings. We hypothesize that this is due to the RT-X dataset mixture not containing many observations from wrist-mounted cameras. For example, less than $10\%$ of the BRIDGE dataset \cite{ebert2021bridge} contains wrist camera observations. As a result, while the performance gap for environments that were seen in the datasets with wrist camera observations is small, the gap for novel environments is greater. We believe that this gap will reduce in the future as more large-scale datasets with egocentric camera are released.
\begin{table}[!htb]
    \centering
    \begin{tabular}{c|c|c|c}
    \toprule
    \textbf{Datasets} & \textbf{Cluttered Grasp} & \textbf{Cluttered Grasp Novel} & \textbf{Toy Kitchen}\\
    \bottomrule
    Wrist + Third & 80\% & 55\% & 80\% \\
    Wrist-only & 75\% & 40\% & 25\% \\
    \bottomrule
    \end{tabular}
     \caption{\textbf{Egocentric Manipulation with Third-person Cotraining.} Policies co-trained with third-person observations have a $42\%$ higher success rate on average than wrist-only policies.}
    \label{table:thirdperson}
\end{table}
\newpage

\subsection{Evaluation Tasks}
Figure~\ref{fig:navmaniptasks} depicts the navigation and manipulation tasks used for evaluation. For manipulation, we evaluate our policies on the two-object grasp, cluttered grasp, toy kitchen, cluttered grasp novel, and shelf manipulation tasks (from left to right). For each environment, we roll out the policy $20$ times to account for higher variances in success when interacting with multiple objects. For navigation, we evaluate our policies on the LoCoBot, Unitree Go-1 and ClearPath Jackal (from left to right). For each embodiment, we average the success rate over $2$ different environments: the office lobby and the kitchen.

\subsection{Additional Discussion}
\subsubsection{Scaling Experiments}
Tables \ref{table:scalingmanip} and \ref{table:scalingnav} show results for models with different numbers of parameters on the manipulation and navigation evaluation tasks. Interestingly, we see a clear positive correlation between model capacity and performance. The 27M param model achieves an average of $33\%$ success on manipulation and $44\%$ on navigation. Meanwhile, the $186M$ param model achieves averages of $68\%$ and $55\%$ success on manipulation and navigation respectively. We hypothesize that the visual diversity across manipulation and navigation contributes to this trend. We record the hyperparameters for the policy architectures we used sorted by parameter count in Table~\ref{table:params}. Encoder represents the size of the EfficientNet image encoder.

\begin{table}[!htb]
    \centering
    \begin{tabular}{c|c|c|c|c}
    \toprule
    \textbf{Datasets} & \textbf{U-Net dims} &  \textbf{Layers} &  \textbf{Encoder} &  \textbf{Embedding Size}\\    
    \bottomrule
    27M  & 128, 256, 512 & 4 & b3 & 1024 \\
    56M  & 128, 256, 512, 1024 & 8,8,8 & b0 & 1024 \\
    66M  & 128, 256, 512, 1024 & 8,8,8 & b0 & 2048 \\
    112M  & 128, 256, 512, 1024 & 8,8,8 & b0 & 2048 \\
    176M  & 128, 256, 512, 1024 & 8,8,8 & b5 & 1024 \\
    186M  & 128, 256, 512, 1024 & 8,8,8 & b5 & 2048 \\
    \bottomrule
    \end{tabular}
     \caption{\textbf{Architectures by Parameter Count.} We record the policies architectures we used for our parameter scaling experiments.}
    \label{table:params}
\end{table}

\subsubsection{Discretization Experiments}
Table~\ref{table:discretization} records the results of training with discretization on various manipulation tasks. Similar to the scheme proposed in RT-1 \cite{brohan2023rt1}, we discretize each action dimension uniformly into $256$ bins. For the model capacities we trained on, we find that our model performs poorly with this discretization scheme, with only the 180M parameter model having any success. Navigation results for discretization are not included in this table because they are flat and cause numerous collisions. We believe that a large model capacity is essential for models with this discretization scheme to perform reasonably. While we were unable to further scale our discretized models due to computation constraints, we believe that performance will increase if the number of parameters are appropriately scaled.

\pagebreak
\newpage
\begin{table*}[!htb]
    \centering
    \begin{tabular}{c|c|c|c|c|c}
    \toprule
    \textbf{Datasets} & \textbf{Two-Object Grasp} & \textbf{Cluttered Grasp} & \textbf{Novel Cluttered Grasp} & \textbf{Toy Kitchen} & \textbf{Shelf Manipulation} \\
    \bottomrule
    GNM + M3ED + Driving + Manip & $85\%$ & $80\%$ & $55\%$ & $80\%$ & $60\%$ \\
    GNM + Driving + Manip  & $80\%$ &  $80\%$ & $50\%$ & $80\%$ & $65\%$\\
    GNM + M3ED + Manip  &  $80\%$ & $70\%$ & $55\%$ & $80\%$ &  $50\%$ \\
    M3ED + Driving + Manip  & $65\%$ & $55\%$ & $25\%$ & $70\%$ & $35\%$ \\
    GNM + Manip &  $80\%$ &  $75\%$ & $50\%$ & $65\%$ & $50\%$\\
    Manip-only &  $70\%$ &  $65\%$&  $20\%$ & $70\%$ & $30\%$ \\
    \bottomrule
    \end{tabular}
     \caption{\textbf{Manipulation Evaluations.} By aligning action coordinate frames, training on navigation and driving datasets results in a $19\%$ improvement across five challenging tabletop manipulation tasks.}
    \label{table:manipeval}
\end{table*}

\begin{table*}[!htb]
    \centering
    \begin{tabular}{c|c|c|c|c|c|c|c|c|c|c|c|c}
    \hline
     & \multicolumn{4}{|c|}{\textbf{Locobot}} & \multicolumn{4}{|c|}{\textbf{DJI Tello}} & \multicolumn{4}{|c}{\textbf{Unitree Go-1}} \\
    \toprule
     \textbf{Datasets}& \multicolumn{2}{|c|}{\textbf{\BAIR Lobby}} & 
     
     \multicolumn{2}{|c|}{\textbf{Kitchen}} & \multicolumn{2}{|c|}{\textbf{\BAIR Lobby}} & \multicolumn{2}{|c|}{\textbf{Kitchen}} & \multicolumn{2}{|c|}{\textbf{\BAIR Lobby}} & \multicolumn{2}{|c}{\textbf{Kitchen}}  \\
     \bottomrule
     & Success & Coll. & Success & Coll. & Success & Coll. & Success & Coll. & Success & Coll. & Success & Coll.\\
    \bottomrule
    GNM + M3ED + BDD + Manip & 85\% & 0 & 97\% & 0 & 40\% & 1 & 57\% & 0.33 & 53\% & 1 &  35\% & 1\\
    GNM + Driving + Manip  & 90\% & 0 & 63\% & 0.33 & 40\% & 1 & 40\% & 1 & 45\% & 1 & 50\% & 0.5 \\
    GNM + M3ED + Manip  &  85\% & 0 & 47\% & 1 & 50\% & 1 & 40\% & 0.5 & 55\% & 1 & 23\% & 1\\
    M3ED + BDD + Manip  & 5\% & 0 & 30\% & 0 & 10\% & 0 & 40\% & 1 & 0\% & 0 & 5\% & 0.5 \\ 
    GNM + Manip &  83\% &  0.33 &  87\% & 0 & 80\%  & 0.5 & 40\% & 1 & 77\% & 1 & 50\% & 0.5 \\
    GNM-only &  70\% &  0 &  70\% &  0 &  65\% & 0.5 & 80\% & 1 & 40\% & 1 & 53\% & 1 \\
    \bottomrule
    \end{tabular}
     \caption{\textbf{Navigation Evaluations.} Across three different robots in challenging indoor and outdoor environments, adding manipulation datasets leads to $13\%$ improvement in navigation performance.}
    \label{table:naveval}
\end{table*}

\label{appendix:scale}
\begin{table*}[!htb]
    \centering
    \begin{tabular}{c|c|c|c|c|c}
    \toprule
    \textbf{Datasets} & \textbf{Two-Object Grasp} & \textbf{Cluttered Grasp} & \textbf{Cluttered Grasp OOD} & \textbf{Toy Kitchen}  & \textbf{Cluttered Grasp} \\
    \bottomrule
    27M Params & $50\%$ & $45\%$ & $20\%$ & $40\%$ & $10\%$ \\
    56M Params & $55\%$ & $45\%$ & $15\%$ & $40\%$ & $20\%$ \\
    66M Params & $75\%$ & $55\%$ & $20\%$ & $70\%$ & $45\%$ \\
    112M Params & $85\%$ & $60\%$ & $45\%$ & $70\%$ & $40\%$ \\
    186M Params & $80\%$ & $80\%$ & $50\%$ & $80\%$ & $50\%$ \\
    \bottomrule
    \end{tabular}
     \caption{\textbf{Manipulation Scaling} Policies with 186M parameters have a $11\%$ higher success rate on average than policies with 27M parameters.}
    \label{table:scalingmanip}
\end{table*}

\begin{table*}[!htb]
    \centering
    \begin{tabular}{c|c|c|c|c|c|c|c|c|c|c|c|c}
    \hline
     & \multicolumn{4}{c}{\textbf{Locobot}} & \multicolumn{4}{c}{\textbf{DJI Tello}} & \multicolumn{4}{c}{\textbf{Unitree Go-1}} \\
    \toprule
      \textbf{Num. Parameters} & \multicolumn{2}{|c|}{\textbf{\BAIR Lobby}} & \multicolumn{2}{|c|}{\textbf{Kitchen}} & \multicolumn{2}{|c|}{\textbf{\BAIR Lobby}} & \multicolumn{2}{|c|}{\textbf{Kitchen}} & \multicolumn{2}{|c|}{\textbf{\BAIR Lobby}} & \multicolumn{2}{c}{\textbf{Kitchen}}  \\
    \bottomrule
    & Success & Coll. & Success & Coll. & Success & Coll. & Success & Coll. & Success & Coll. & Success & Coll. \\
    \bottomrule
    27M Params & 68\% & 0.33 & 95\% & 0 & 33\% & 0 & 18\% & 1 & 30\% & 1 & 20\% & 1\\
    56M Params & 77\% & 0.67 & 53\% & 0.67 & 28\% & 1 & 40\% & 1 & 25\% & 1 & 20\% & 0.5\\
    66M Params & 87\% & 0.33 & 50\% & 0.33 &  65\% & 1 & 48\% & 1 & 30\% & 1 & 40\% & 1\\
    112M Params & 90\% & 0.33 & 85\% & 0.33 & 65\% & 0.5 & 50\% & 1 & 58\% & 0.5 & 73\% & 0.5\\
    176M Params & 83\% & 0.33 & 73\% & 0.67 & 68\% & 0.5 & 60\% & 0 & 40\% & 0.5 & 53\% & 1\\
    186M Params & 90\% & 0 & 63\% & 0.33 & 40\% & 1 & 40\% & 1 & 45\% & 1 & 50\% & 0.5 \\
    \bottomrule
    \end{tabular}
     \caption{\textbf{Navigation Scaling.} Policies with 186M parameters have a $35\%$ higher success rate on average than policies with 27M parameters on manipulation tasks.}
    \label{table:scalingnav}
\end{table*}

\begin{table*}[!htb]
    \centering
    \begin{tabular}{c|c|c|c|c}
    \toprule
    \textbf{Datasets} & \textbf{Two-Object Grasp} & \textbf{Cluttered Grasp} & \textbf{Cluttered Grasp OOD} \\
    \bottomrule
    40M & $0\%$ & $0\%$ & $0\%$ \\
    100M & $0\%$ & $0\%$ & $0\%$ \\
    180M  & $10\%$ & $5\%$& $5\%$\\
    \bottomrule
    \end{tabular}
     \caption{\textbf{Discretization Experiments.} Policies trained with a discretization head and cross-entropy loss under 100M parameters fail to have nonzero success.}
    \label{table:discretization}
\end{table*}

\begin{table*}[!htb]
    \centering
    \begin{tabular}{c|c|c|c|c}
    \toprule
    \textbf{Datasets} & \textbf{Two-Object Grasp} & \textbf{Cluttered Grasp} & \textbf{Novel Cluttered Grasp} & \textbf{Toy Kitchen} \\
    \bottomrule
    Polybot + RT-X + Sacson &  $80\%$ & $70\%$ & $35\%$ &  $70\%$ \\
    Polybot + RT-X + Go Stanford & $65\%$ & $6\%$ &  $35\%$ & $75\%$ \\
    Polybot + RT-X + Tartan & $25\%$ &  $10\%$  & $10\%$ & $0\%$ \\
    Polybot + RT-X + Recon &  $50\%$ &  $50\%$ & $20\%$ & $60\%$ \\
    Polybot + RT-X + Cory Hall & $55\%$ & $5\%$ & $20\%$ & $70\%$ \\
    Polybot + RT-X + Seattle &  $20\%$  & $5\%$ &  $10\%$ & $10\%$ \\
    Polybot + RT-X + Scand &  $55\%$ & $50\%$ &  $20\%$ & $45\%$ \\
    \bottomrule
    \end{tabular}
    \caption{\textbf{GNM Ablations.} Manipulation policies co-trained with indoor and outdoor navigation data on sidewalks perform better than policies co-trained on outdoor navigation data in off-road or trail-like environments.} 
    \label{table:gnmablations}
\end{table*}

\pagebreak
\newpage